\documentclass[10pt, conference, compsocconf]{IEEEtran}
\ifCLASSINFOpdf
\else
\fi
\usepackage{amssymb}
\usepackage{amsmath}                            
\usepackage{graphicx}                    
\usepackage{algorithmicx, algpseudocode}  
\usepackage{algorithm}    
\usepackage{dsfont}
\usepackage{pifont}   
\usepackage{cite}

\newcommand{\xmark}{\ding{55}}%
\newcommand{\cmark}{\ding{52}}%
\makeatletter

\begin{document}
\title{Longitudinal Face Aging in the Wild - Recent Deep Learning Approaches}

 \author{\IEEEauthorblockN{Chi Nhan Duong}
 \IEEEauthorblockA{Concordia University\\
 Montreal, Quebec, Canada\\
 Email: c\_duon@encs.concordia.ca}
 \\ \\ 
 \IEEEauthorblockN{Kha Gia Quach}
 \IEEEauthorblockA{Concordia University\\
 Montreal, Quebec, Canada\\
 Email: k\_q@encs.concordia.ca}
 \and
 \IEEEauthorblockN{Khoa Luu}
 \IEEEauthorblockA{CyLab Biometrics Center\\ 
 Dept. of Electrical and Computer Engineering\\
 Carnegie Mellon University Pittsburgh, PA, USA\\
 Email: kluu@andrew.cmu.edu}
 \\
 \centering
 \IEEEauthorblockN{Tien D. Bui}
 \IEEEauthorblockA{Concordia University\\
 Montreal, Quebec, Canada\\
 Email: bui@encs.concordia.ca}
 }

\maketitle

\begin{abstract}
Face Aging has raised considerable attentions and interest from the computer vision community in recent years. Numerous approaches ranging from purely image processing techniques to deep learning structures have been proposed in literature. 
In this paper, we aim to give a review of recent developments of modern deep learning based approaches, i.e. Deep Generative Models, for Face Aging task.
Their structures, formulation, learning algorithms as well as synthesized results are also provided with systematic discussions.
Moreover, the aging databases used in most methods to learn the aging process are also reviewed. 
\end{abstract}

\begin{IEEEkeywords}
Face Aging, Face Age Progression, Deep Generative Models.

\end{IEEEkeywords}

\IEEEpeerreviewmaketitle

\begin{table*}[!t] 
	\small 
	\centering
	\caption{Properties of different aging databases.``In-the-wild'' images are the ones collected from unconstrained real-world conditions.} 
	\label{tab:AgingDatabaseProperties}
	\begin{tabular}{|l|c|c|c|c|c|c|c|}
		\hline
		\textbf{Database} & \textbf{\# Images} & \textbf{\# Subjects} & \textbf{Label type} &  \textbf{Image type} & \textbf{Subject type}&  \textbf{Clean Label} & \textbf{Public}\\  
		\hline \hline
		MORPH - Album 1 \cite{ricanek2006morph} & 1690 & 628 & Years old & Mugshot &  Non-famous & \cmark & \cmark\\				
		MORPH - Album 2 \cite{ricanek2006morph} & 55134 & 13000 & Years old & Mugshot &  non-famous &\cmark & \cmark\\
		\hline \hline
		FG-NET \cite{fgNetData} & 1002 & 82 & Years old & In-the-wild &  Non-famous &\cmark & \cmark\\
		AdienceFaces \cite{levi2015age} & 26580 & 2984 & Age groups & In-the-wild &  Non-famous & \cmark & \cmark\\
		CACD \cite{chen14cross} & 163446 & 2000 & Years old & In-the-wild & Celebrities & \xmark & \cmark\\
		IMDB-WIKI \cite{Rothe-IJCV-2016} & 523051 & 20284 & Years old & In-the-wild &  Celebrities &\xmark & \cmark\\
        
        AgeDB \cite{AgeDB} & 16488 & 568 & Years old & In-the-wild & Celebrities & \cmark & \cmark\\
        \hline
		\hline
		\textbf{AGFW (Ours) \cite{Duong_2016_CVPR}} & \textbf{18685} & \textbf{14185} & \textbf{Age groups} & \textbf{In-the-wild/Mugshot} & \textbf{Non-famous} & \cmark & \cmark\\
        \textbf{AGFW-v2 (Ours)}& \textbf{36325} & \textbf{27688} & \textbf{Age groups} & \textbf{In-the-wild/Mugshot} & \textbf{Non-famous} & \cmark & \cmark\\
		\hline
		
	\end{tabular}
\end{table*}

\section{Introduction}
In recent years, age progression has received considerable interest from the computer vision community. 
Starting from the predominant approaches that require lots of time and professional skills with the support from forensic artists, several breakthroughs have been achieved. Numerous automatic age progression approaches from anthropology theories to deep learning models have been proposed.
In general, the age progression methods can be technically classified into four categories, i.e. modeling, reconstruction, prototyping and deep learning based methods.
The methods in the first three categories usually tend to simulate the aging process of facial features by (1) adopting prior knowledge from anthropometric studies 
; or (2) representing the face geometry and appearance by a set of parameters via conventional models such as Active Appearance Models (AAMs), 3D Morphable Models (3DMM) and manipulate these parameters via learned aging functions. Although they have achieved some inspiring synthesis results, these face representations are still linear and facing lots of limitations in modeling the non-linear aging process.
\begin{figure}[!t]
	\begin{center}
		\includegraphics[width=8.5cm]{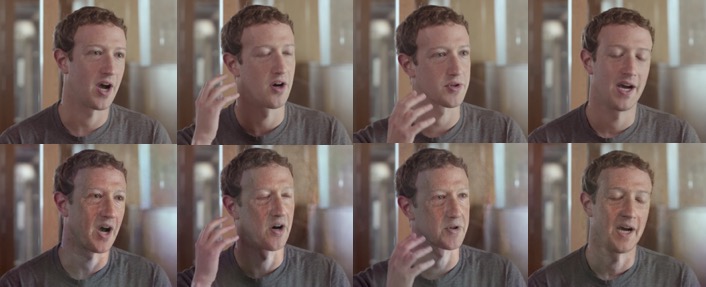}\end{center}
	\caption{Given input images, Age Progression task is to predict the future faces of that subject many years later.}
	\label{fig:Aging_Example_Trump}
\end{figure}

Meanwhile, the fourth category introduces modern approaches with the state-of-the-art \textbf{\textit{Deep Generative Models}} (DGM) for both face modeling and aging embedding process. Since deep learning structures have more capabilities of interpreting and transferring the highly non-linear features of the input signals, they are more suitable for modeling the human aging process. As a result, superior synthesized facial images \cite{Duong_2016_CVPR,Duong_2017_ICCV}, \cite{duong2017learning,Zhang_2017_CVPR,wang2016recurrent} can be generated. 
Inspired by these state-of-the-art results, in this paper, we aim to provide a review of recent developments for face age progression. 
Both \textit{structures and formulations} of several Deep Generative Models, i.e. Restricted Boltzmann Machines (RBM), Deep Boltzmann Machines (DBM), and Generative Adversarial Network (GAN), as well as \textit{the way they are adopted to age progression problem} will be presented. 
Moreover, several common face aging databases are also reviewed.

\begin{figure}[!t]
	\begin{center}
		\includegraphics[width=8.5cm]{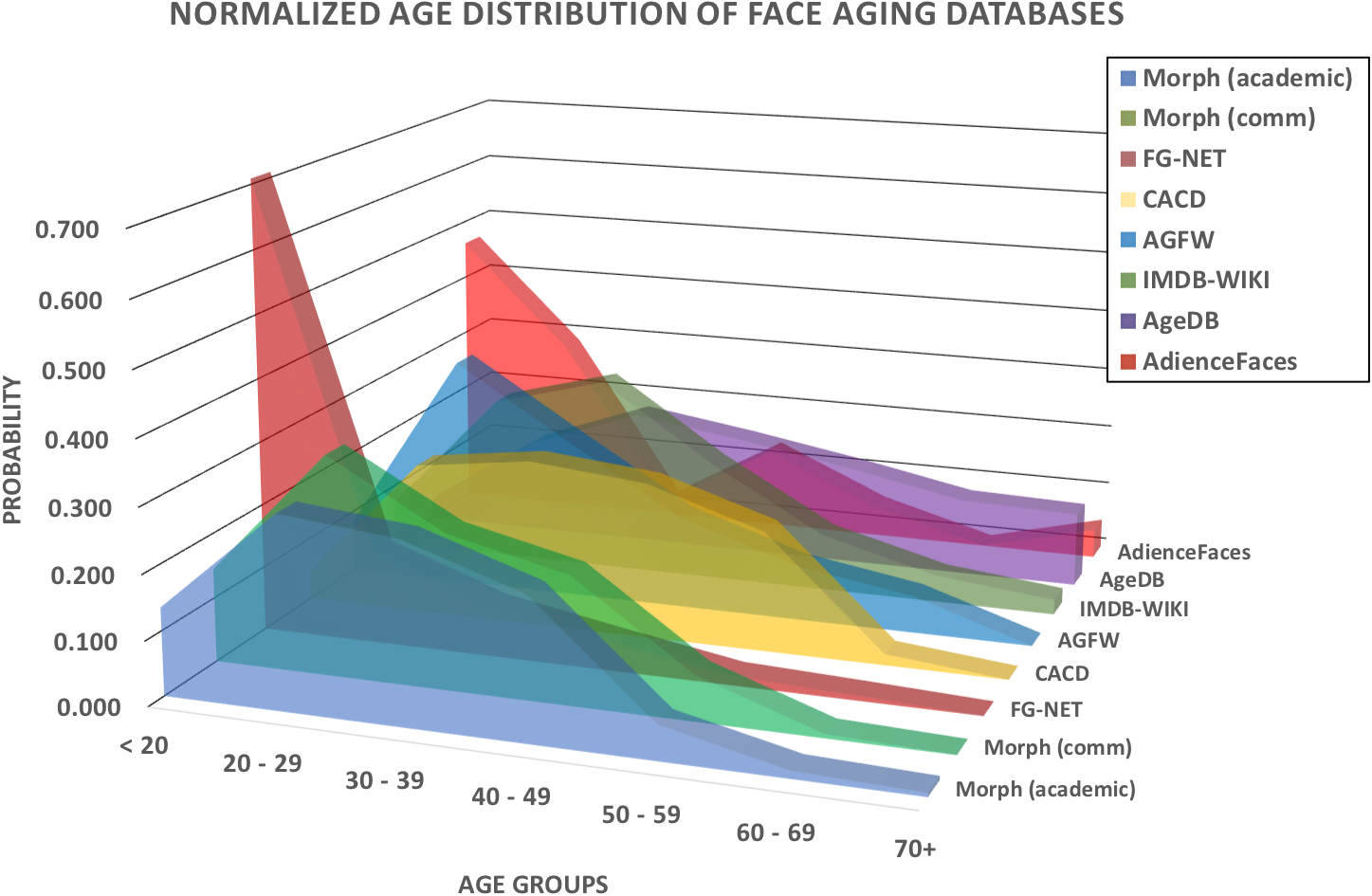}\end{center}
	\caption{Normalized Age Distribution of Face Aging Databases}
	\label{fig:FG_DB_SUM}
\end{figure}

\section{Face Aging Databases} \label{sec:Database}

Database collection for face aging is also a challenging problem.
There are several requirements during the collecting process. Not only should each subject have images at different ages, but also the covered age range should be large.
Therefore, face aging databases are still limited in terms of age labels and the number of available databases. 
The characteristics and age distributions of several current existing face aging databases are summarized in Table \ref{tab:AgingDatabaseProperties} and Fig. \ref{fig:FG_DB_SUM}.
\begin{table*}[!t] \label{tab:ConventionalAP}
	\small
	\centering
	\caption{A summary of Conventional Approaches for Age Progression.} 
	\label{tb:ConventionalAPApproaches} 
	\begin{tabular}{|l|c|c|c|l|}
		\hline
		\textbf{Method} 
		& \textbf{Approach} 
		& \begin{tabular}[l]{@{}l@{}} \textbf{Represent}\textbf{ation}\end{tabular} 
		& \textbf{Architecture} 
		& \textbf{Summary} \\
		\hline \hline
		\begin{tabular}[l]{@{}l@{}} Lanitis et al. \\2002 \cite{lanitis2002toward} \end{tabular}& Model based & AAMs&\xmark & \begin{tabular}[l]{@{}l@{}} Model generic and specific aging processes\\ four types of aging functions \end{tabular}\\
		\hline
		\begin{tabular}[l]{@{}l@{}}Pattersons et al. \\ 2006\cite{patterson2006automatic} \end{tabular}& Model based &AAMs & \xmark & \begin{tabular}[l]{@{}l@{}} Learning Effects of morphological changes \\  More efforts on the adult aging stage\end{tabular}\\
		\hline

		\begin{tabular}[l]{@{}l@{}}Luu et al. \\2009\cite{luu2009Automatic} \end{tabular}&Model based &AAMs &\xmark & Incorporated familial facial cues\\
		\hline
		\begin{tabular}[l]{@{}l@{}}Geng et al. \\2007 \cite{geng2007automatic} \end{tabular}&Model based & Aging Patterns & Aging Pattern Subspace & Grammatical face model \\
		\hline
		\begin{tabular}[l]{@{}l@{}}Tsai et al. \\2014 \cite{tsai2014human}  \end{tabular}&Model based &Aging Patterns & Aging Pattern Subspace & \begin{tabular}[l]{@{}l@{}}Guidance faces according to subject's feature \end{tabular}\\
		\hline
		\begin{tabular}[l]{@{}l@{}}Suo et al. \\2010 \cite{suo2010compositional } \end{tabular}&Model based & Part-based & And-Or-Graph & Markov Chain on Parse Graphs \\
		\hline
		\begin{tabular}[l]{@{}l@{}}Suo et al. \\2012 \cite{suo2012concatenational } \end{tabular}& Model based& Part-based & And-Or-Graph & Composition of short-term graph evolution\\
		\hline
		\begin{tabular}[l]{@{}l@{}} Kemelmacher-\\Shlizerman et al. \\2014 \cite{kemelmacher2014illumination} \end{tabular}& Prototype & Image pixel & \xmark & \begin{tabular}[l]{@{}l@{}} Illumination normalization and subspace \\alignment before transferring difference \\ between prototypes. \end{tabular}\\
		\hline
		\begin{tabular}[l]{@{}l@{}}	Shu et al.\\2015  \cite{Shu_2015_ICCV} \end{tabular}& Reconstructing& Sparse Representation & Coupled dictionaries& \begin{tabular}[l]{@{}l@{}}Dictionary Learning with personality-aware \\coupled reconstruction loss\end{tabular}\\
		\hline
		\begin{tabular}[l]{@{}l@{}}Yang et al. \\2016 \cite{yang2016face} \end{tabular}& Reconstructing&Sparse Representation & Hidden Factor Analysis &  \begin{tabular}[l]{@{}l@{}}Sparse reconstruction with age-specific \\feature \end{tabular}\\
		\hline
	\end{tabular}
\end{table*}
Further than these databases, a large-scale in-the-wild dataset, named \textbf{AGing Face in-the-Wild (AGFW)}, was also introduced in our work \cite{Duong_2016_CVPR} with 18,685 facial images with individual ages sampled ranging from 10 to 64. 
In this database, images are divided into 11 age groups with the span of 5 years, each group contains 1,700 images on average. This database is then extended to \textbf{AGFW-v2} with double scale, i.e. 36,325 images with an average of 3,300 images per age group.

\section{Conventional Approaches} \label{sec:ConventionalApproaches}
In this section, we provide a brief review of conventional age progression approaches including modeling, prototyping, and reconstructing based approaches. Their properties are also summarized in Table \ref{tab:ConventionalAP}.
\subsection{Modeling-based approach}
Modeling-based approach is among the earliest categories presented for face age progression.
These methods usually exploit some kinds of appearance models, i.e. Active Appearance Models (AAM), 3D Morphable Models (3DMM), to represent the  shapes and texture of the input face by a set of parameters. Then the aging process is simulated by learning some aging functions from the relationship of the parameter sets of different age groups.
In particular, Pattersons et al. \cite{patterson2006automatic} and Lanitis et al. \cite{lanitis2002toward} employed a set of Active Appearance Models (AAMs) parameters with four aging functions to model both the general and the specific aging processes.
Four variations of aging functions were introduced: Global Aging Function, Appearance Specific Aging Function (ASA), Weighted Appearance Aging Function (WAA), and Weighted Person Specific Aging Function (WSA).
Also by employing AAMs during the modeling step, Luu et al. \cite{luu2009Automatic} later incorporated familial facial cues to the process of face age progression.

Another direction of modeling was proposed in \cite{geng2007automatic} with a definition of AGing pattErn Subspace (AGES). In this approach, the authors construct a representative subspace for aging patterns as a chronological sequence of face images. Then given an image, the proper aging pattern is determined by the projection in this subspace that produces smallest reconstruction error. Finally, the synthesized result at a target age is obtained by the reconstructed faces corresponding to that age position in the subspace.
Tsai et al. \cite{tsai2014human} then enhanced the AGES using guidance faces corresponding to the subject's characteristics to produce more stable results. 
Suo et al. \cite{suo2010compositional, suo2012concatenational} introduced the three-layer And-Or Graph (AOG) of smaller parts, i.e. eyes, nose, mouth, etc., to model a face. Then, the face aging process was learned for each part using a Markov chain.

\begin{figure}[!t]
	\begin{center}
		\includegraphics[width=7.4cm]{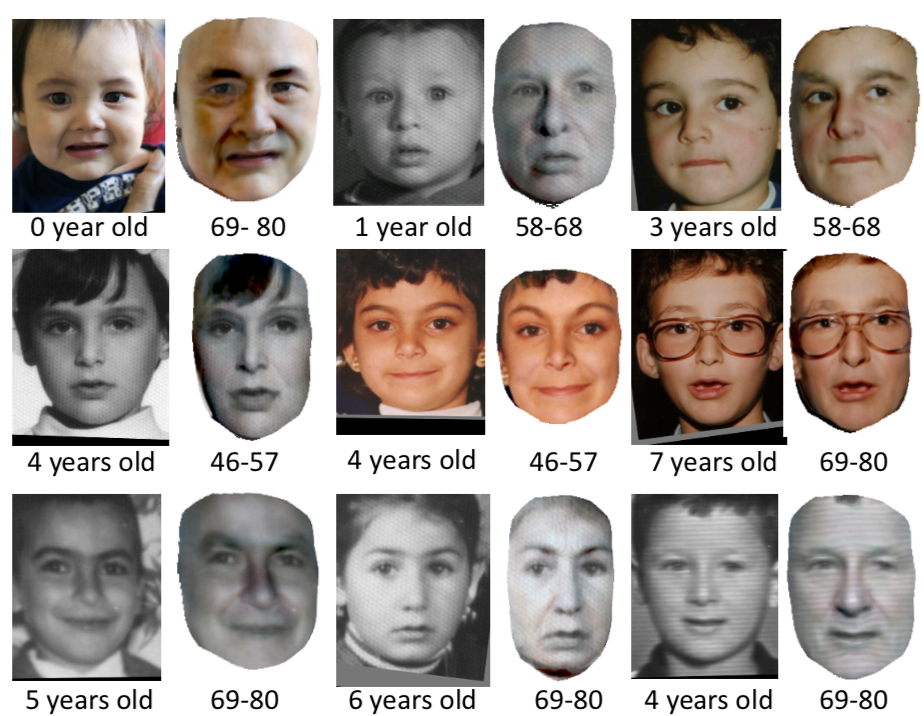}\end{center}
	\caption{Examples of age-progressed faces obtained by Illumination-Aware Age Progression approach\cite{kemelmacher2014illumination}.}
	\label{fig:IAAP_results}
\end{figure}

\subsection{Prototyping approach}
The main idea of the methods in this category is to predefine some types of \textit{aging prototypes} and transfer the difference between these prototypes to produce synthesized face images. Usually, the aging prototypes are defined by the average faces of all age groups  \cite{rowland1995manipulating}. Then, input face image can be progressed to the target age by incorporating the differences between the prototypes of two age groups \cite{burt1995perception}. Notice that this approach requires a good alignment between faces in order to produce plausible results.
Kemelmacher-Shlizerman et al. \cite{kemelmacher2014illumination} then proposed to construct high quality average prototypes from a large-scale set of images. Sharper average faces are obtained via the collection flow method introduced in \cite{liu2011sift} to align and normalize all the images in one age group. Then illumination normalization and subspace alignment technique are employed to handle images with various lighting conditions.
Figure \ref{fig:IAAP_results} illustrates the results obtained in \cite{kemelmacher2014illumination}.

\subsection{Reconstructing-based approach}
Rather than constructing aging prototypes for each age group, the reconstructing-based methods focus on constructing the ``aging basis'' for each age group and model aging faces by the combination of these bases. Dictionary learning techniques are usually employed for this type of approach. 
Shu et al. \cite{Shu_2015_ICCV} proposed to use the aging coupled dictionaries (CDL) to model personalized aging patterns by preserving personalized facial features. The dictionaries are learned using face pairs from neighboring age groups via a personality-aware coupled reconstruction loss. 
Yang et al. \cite{yang2016face} represented person-specific and age-specific factors independently using sparse representation hidden factor analysis (HFA). Since only age-specific gradually changes over time, the age factor is transformed to the target age group via sparse reconstruction and then combined with the identity factor to achieve the aged face.

\section{Deep Generative Models for Face Aging} \label{sec:DGMApproaches}
In this section, we firstly provide an overview of the structures and formulations of the common Deep Generative Models before going through the age progression techniques developed from these structures.

\subsection{From Linear Models to Deep Structures}
Compared to linear models such as AAMs and 3DMM, deep structures have gained significant attention as one of the emerging research topics in both representing higher-level data features and learning the distribution of observations. For example, being designed following the concepts from Probabilistic Graphical Models (PGM),  the RBM-based models organize their non-linear latent variables in multiple connected layers with an energy function such that each layer can learn a different factor to represent the data variations. 
This section introduces the structures, formulations of several Deep Generative Models including RBM, Deep Boltzmann Machines (DBM), Generative Adversarial Networks (GANs).
\subsubsection{Restricted Boltzmann Machines (RBM)\cite{hinton2002training}} are undirected graphical models consisting two layers of stochastic units, i.e. visible
 $\mathbf{v}$ and hidden units  $\mathbf{h}$. This is a simplified version of Boltzmann Machines where no intra connections between units in the same layer is created. RBM structure is a bipartite graph where visible and hidden units are pairwise conditionally independent. 
Given a binary state of $\mathbf{v,h}$, 
the energy of RBM and the joint distribution of visible and hidden units can be computed as
\small
\begin{equation} \label{eqn:RBM}
\begin{split}
-E(\mathbf{v,h}) &= \mathbf{v}^T\mathbf{Wh}+\mathbf{b}^T\mathbf{v}+\mathbf{a}^T\mathbf{h}\\
P(\mathbf{v,h};\mathbf{\theta}) &= \frac{1}{Z(\mathbf{\theta})} \exp \{-E(\mathbf{v,h})\}
\end{split}
\normalsize
\end{equation}
where $\mathbf{\theta}=\{\mathbf{W,b,a}\}$ denotes the parameter set of RBM including the connection weights and the biases of visible and hidden units, respectively.
The conditional probabilities RBM structure can be computed as $p(h_j = 1 | \mathbf{v}) = \sigma (\sum_i v_i w_{ij} + a_j)$ and $p(v_i=1|\mathbf{h}) = \sigma(\sum_j h_j w_{ij} + b_i)$ where $\sigma(\cdot)$ is the logistic function.

\begin{figure}[!t]
	\begin{center}
		\includegraphics[width=8.5cm]{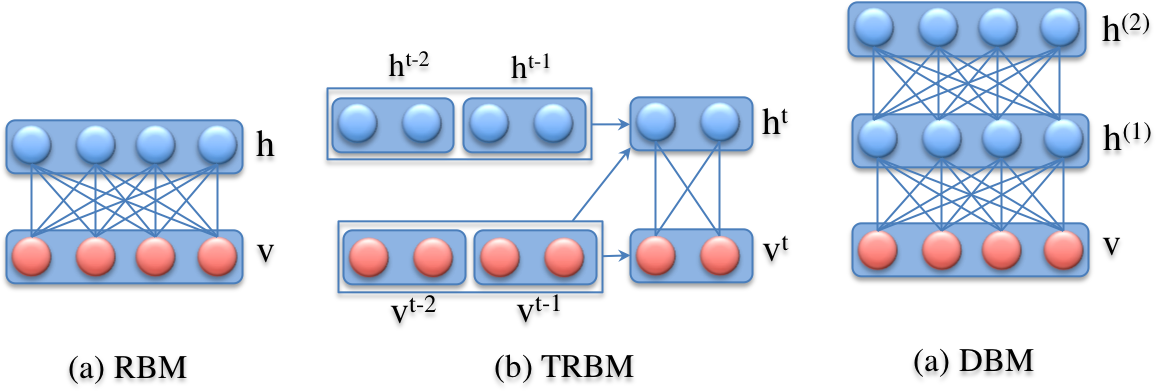}\end{center}
	\caption{Structures of (a) RBM, (b) TRBM and (c) DBM.}
	\label{fig:RBM_TRBM_DBM}
\end{figure}

In the original RBM, both visible and hidden units are binary. 
To make it more powerful and be able to deal with real-valued data, an extension of RBM, named \textit{Gaussian Restricted Boltzmann Machine}, is introduced in \cite{krizhevsky2009learning}.
In Gaussian RBM, the visible units are assumed to have values in $[-\infty, \infty]$ and normally distributed with mean $b_i$ and variance $\sigma_i^2$. Another extension of RBM is Temporal Restricted Boltzmann Machines (TRBM) \cite{sutskever2007learning} which was designed to model complex time-series structure. The structure of TRBM is shown in Fig. \ref{fig:RBM_TRBM_DBM} (b). The major difference between the original RBM and TRBM is the directed connections from both visible and hidden units of previous states to the current states. 
With these new connections, the short history of their activations can act as ``memory'' and is able to contribute to the inference step of current states of visible units. 

\subsubsection{Deep  Boltzmann Machines (DBM)}
As an extension of RBM with more than one hidden layer, the structure of DBM contains several RBMs organized in layers. Thanks to this structure, the hidden units in higher layer can learn more complicated correlations of features captured in lower layer. Another interesting point of DBM is that these higher representations can be built from the training data in an unsupervised fashion. Unlike other models such as Deep Belief Network \cite{hinton2006reducing} or Deep Autoencoders \cite{bengio2009learning}, all connections between units in two consecutive layers are undirected. As a result, each unit receives both bottom-up and top-down information and, therefore, can better propagate uncertainty during the inference process.

Let $\{\mathbf{h}^{(1)},\mathbf{h}^{(2)}\}$ be the set of units in two hidden layers, the energy of the state $\{\mathbf{v}, \mathbf{h}^{(1)},\mathbf{h}^{(2)}\}$ is given as follows.
\small
\begin{equation} \label{eqnEnergyDBM}
\begin{split}
-E(\mathbf{v}, \mathbf{h}^{(1)},\mathbf{h}^{(2)};\boldsymbol{\theta}) = &\mathbf{v}^\top\mathbf{W}^{(1)}\mathbf{h}^{(1)} 
+ \mathbf{h}^{(1)\top}\mathbf{W}^{(2)}\mathbf{h}^{(2)}
\end{split}
\end{equation}
\normalsize
where $\boldsymbol{\theta} = \{\mathbf{W}^{(1)},\mathbf{W}^{(2)}\}$ are the weights of visible-to-hidden and hidden-to-hidden connections. Notice that the bias terms for visible and hidden units are ignored in Eqn. (\ref{eqnEnergyDBM}) for simplifying the representation.
Exploiting the advantages of DBM, Deep Appearance Models (DAM) \cite{duong2015beyond} and Robust Deep Appearance Models (RDAM) \cite{quach2016robust} have been introduced and proven to be superior to other classical models such as AAMs in inferencing a representation for new face images under various challenging conditions. 

\subsubsection{Generative Adversarial Networks (GAN)} \label{sec:GANStructure}
In order to avoid the intractable Markov chain sampling during the training stage of RBM, Goodfellow et al. \cite{goodfellow2014generative} borrowed the idea from adversarial system to design their Generative Adversarial Networks (GAN). The intuition behind this approach is to set up a game between \textit{generator} and \textit{discriminator}.
On one hand, the discriminator learns to determine whether given data are from the generator or real samples. On the other hand, the generator learns how to fool the discriminator by its generated samples. This game continues as the learning process takes place. The learning process will stop at a point that the discriminator can't distinguish between real data and the ones produced by the generator. This is also an indication that the generator has already learned the distribution of input data.
Formally, let $\mathbf{x}$ be the input data, $p_g$ be the distribution learned from generator, and $p_z(\mathbf{z})$ be the prior distribution of variable $\mathbf{z}$. Then GAN is defined by two neural networks representing two differentiable functions for the generator $G(\mathbf{z}, \theta_g): \mathbf{z} \mapsto \mathbf{x}$ and discriminator $D(\mathbf{x}, \theta_d): \mathbf{x} \mapsto y$
where $y$ denotes the probability that $\mathbf{x}$ comes from the data distribution rather than $p_g$; $\theta_g$ and $\theta_d$ are the parameters of the CNNs representing $G$ and $D$, respectively. The training process is then formulated as maximizing the probability $D(\mathbf{x})$ while minimizing $\log \left(1 - D(G(\mathbf{z}))\right)$:
\small
\begin{equation} \label{eqn:GANformulation}
\begin{split}
\min_G \max_D V(D,G) = & \mathbb{E}_{\mathbf{x}\sim p_{data}(\mathbf{x})}\left[\log D(\mathbf{x})\right] \\
& + \mathbb{E}_{\mathbf{z}\sim p_{\mathbf{z}}(\mathbf{z})}\left[\log \left( 1 - D(G(\mathbf{z}))\right)\right]
\end{split}
\end{equation}
\normalsize
In original GAN, the use of fully connected neural network for its generator makes it very hard to generate high-resolution face images. 
Then numerous extensions of GAN focusing on different aspects of this structure have been proposed in literature such as Laplacian pyramid Generative Adversarial Networks (LAPGAN) \cite{denton2015deep},  Deep Convolutional Generative Adversarial Networks (DCGAN) \cite{radford2015unsupervised}, Info-GAN \cite{chen2016infogan}, Wasserstein GAN \cite{arjovsky2017wasserstein}.

\begin{table*}[!t]
	\small
	\centering
	\caption{Properties of Deep Generative Model Approaches for Age Progression. Deep Learning (DL), Log-Likelihood (LL), Inverse Reinforcement Learning (IRL), Probabilistic Graphical Models (PGM), Adversarial (ADV)} 
	\label{tb:ComputerAPApproaches} 
	\begin{tabular}{|l|c|c|c|c|c|c|c|c|}
		\hline
		\textbf{Method} 
		& \textbf{Approach} 
		& \textbf{Architecture} & \begin{tabular}[l]{@{}l@{}} \textbf{Loss}\\ \textbf{Function} \end{tabular}& \begin{tabular}[l]{@{}l@{}} \textbf{Non}\\\textbf{-Linearity} \end{tabular}& \begin{tabular}[l]{@{}l@{}} \textbf{Tractable} \\\textbf{Model} \end{tabular}
		& \begin{tabular}[l]{@{}l@{}}\textbf{Subject}\\\textbf{Dependent}\end{tabular}
		& \begin{tabular}[l]{@{}l@{}}\textbf{Multiple} \\\textbf{Input}\\\textbf{support}\end{tabular}
		\\  
		\hline \hline

		\begin{tabular}[l]{@{}l@{}}Wang et al. 2016 \cite{wang2016recurrent} \end{tabular}& DL
		& RNN & $\ell_2$ & \cmark & \cmark & \xmark & \xmark\\
		\hline
		\begin{tabular}[l]{@{}l@{}}Zhang et al. 2017 \cite{Zhang_2017_CVPR} \end{tabular}& DL
		& GAN & ADV + $\ell_2$ & \cmark & \cmark & \xmark & \xmark\\
		\hline
		\begin{tabular}[l]{@{}l@{}}Antipov et al. 2017 \cite{antipov2017face} \end{tabular}& DL
		& GAN & ADV + $\ell_2$ & \cmark & \cmark & \xmark & \xmark\\
		\hline
		\begin{tabular}[l]{@{}l@{}}Li et al. 2018 \cite{li2018global} \end{tabular}& DL 
		& GAN & \begin{tabular}[l]{@{}l@{}}ADV + $\ell_2$  + ID + Age \end{tabular}& \cmark & \cmark & \xmark & \xmark\\
		\hline \hline
        \begin{tabular}[l]{@{}l@{}}\textbf{Ours, 2016 \cite{Duong_2016_CVPR}} \end{tabular}& \textbf{DL} 
		& \textbf{TRBM} & \textbf{LL} & \cmark & \xmark & \xmark & \xmark\\
		\hline
		\begin{tabular}[l]{@{}l@{}}\textbf{Ours, 2017 \cite{Duong_2017_ICCV}} \end{tabular}& \textbf{DL} 
		& \begin{tabular}[l]{@{}l@{}}\textbf{PGM + CNN }\end{tabular}& \begin{tabular}[l]{@{}l@{}}\textbf{LL} \end{tabular}& \cmark & \cmark & \xmark & \xmark\\
		\hline
		\begin{tabular}[l]{@{}l@{}}\textbf{Ours, 2017 \cite{duong2017learning}} \end{tabular}& \textbf{DL + IRL} 
		& \begin{tabular}[l]{@{}l@{}}\textbf{PGM + CNN }\end{tabular}& \begin{tabular}[l]{@{}l@{}}\textbf{LL} \end{tabular}& \cmark & \cmark & \cmark & \cmark\\
		\hline
	\end{tabular}
\end{table*}

\begin{figure*}[!t]
	\begin{center}
		\includegraphics[width=17cm]{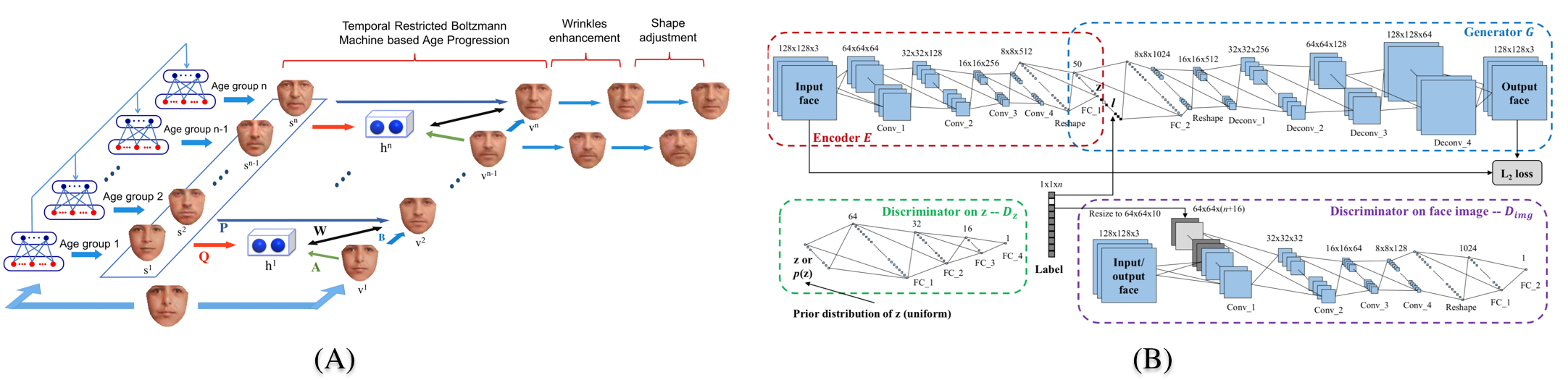}\end{center}
	\caption{Architectures of (A) the TRBM-based model \cite{Duong_2016_CVPR} with post-processing steps, and (B) GAN-based Age Progression model \cite{Zhang_2017_CVPR}.}
	\label{fig:TRBM_GAN}
\end{figure*}

\subsection{Deep Aging Models for Age Progression}
Thanks to the power of Deep Learning models in terms of non-linear variations modeling, many deep learning based age progression approaches have been recently developed and achieved considerable results in face age progression. 
Table \ref{tb:ComputerAPApproaches} summarizes the key features of these deep learning based approaches.

\begin{figure}[!t]
	\begin{center}
		\includegraphics[width=7.5cm]{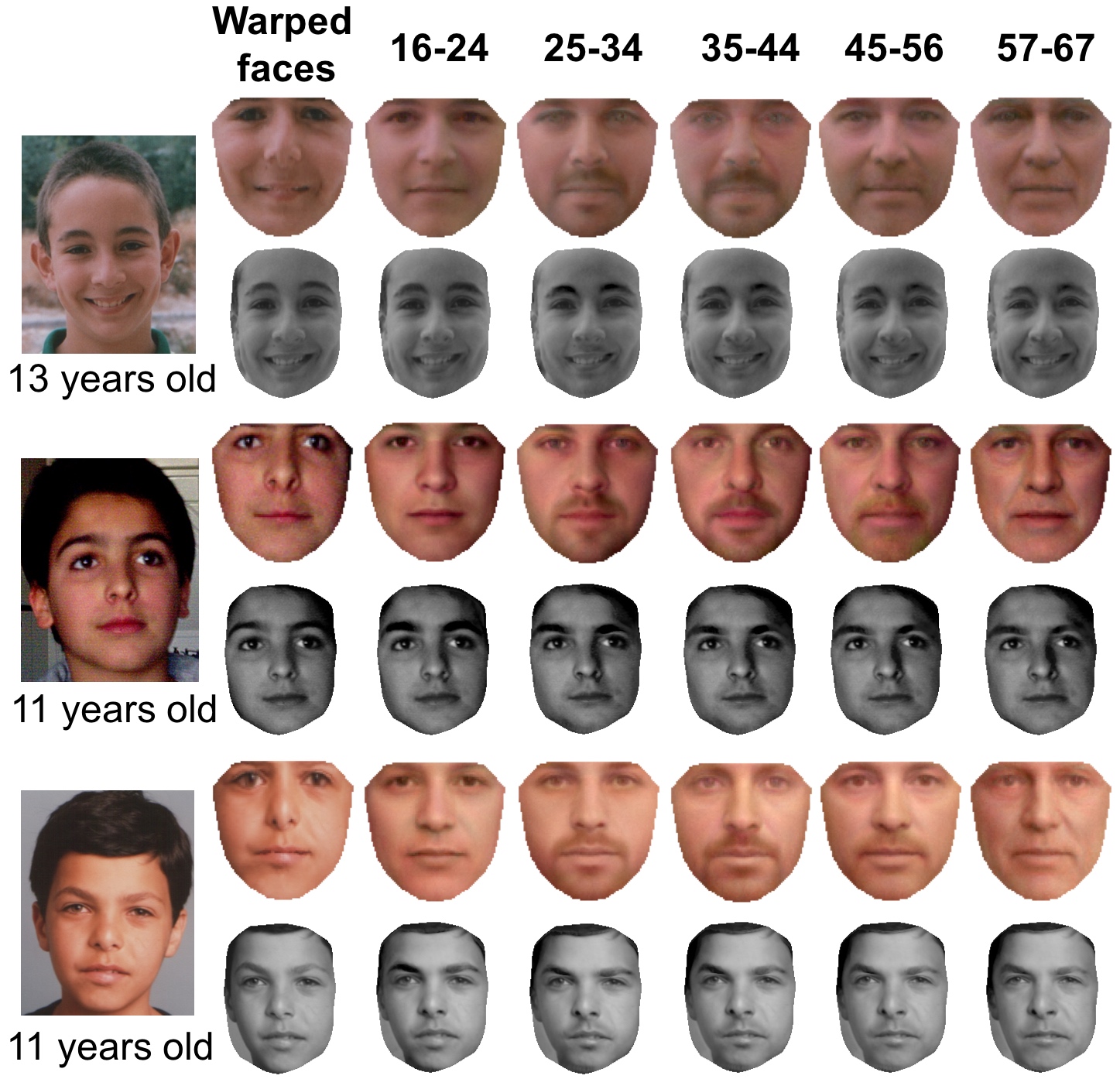}\end{center}
	\caption{A comparison between age-progressed sequence generated by TRBM based model\cite{Duong_2016_CVPR} and IAAP approach\cite{kemelmacher2014illumination}.}
	\label{fig:CompareTRBM_IAAP}
\end{figure}

\noindent
\paragraph{\textbf{TRBM-based model}}
In addition to single face modeling, a TRBM based age progression model is introduced in \cite{Duong_2016_CVPR} to embed the temporal relationship between images in a face sequence. 
By taking the advantages of log-likelihood objective function and avoiding the $\ell_2$ reconstruction error during training, the  model is able to efficiently capture the non-linear aging process and automatically synthesize a series of age-progressed faces in various age ranges with more aging details. This approach has presented a carefully designed architecture with the combination of both RBM and TRBM for age variation modeling and age transformation embedding.
Fig. \ref{fig:TRBM_GAN}(A) illustrates the aging architecture with TRBM proposed in \cite{Duong_2016_CVPR}.
In this approach, the long-term aging development is considered as a composition of short-term changes and can be represented as a sequence of that subject faces in different age groups. After the decomposition, a set of RBMs is employed to model the age variation of each age group as well as the wrinkles presented in the faces of older ages. Then the TRBM based model is constructed to embed the aging transformation between faces of consecutive age groups.
Particularly, keeping similar form of the energy function as original TRBM and RBM , the bias terms are defined as
	\small
\begin{equation}
\begin{split}
b_i^t =& b_i + B_i \mathbf{v}^{t-1} + \sum_l P_{li}\mathbf{s}_l^{<=t}\\
a_j^t =& a_j + A_j \mathbf{v}^{t-1} + \sum_l Q_{lj}\mathbf{s}_l^{<=t}
\end{split}
\end{equation}
\normalsize
where $\{\mathbf{A,B,P,Q} \}$ are the model parameters; and $\mathbf{s}^{<=t} = \{\mathbf{s}^t, \mathbf{s}^{t-1}\}$ denote the reference faces produced by the set of learned RBM. With this structure, both linear and non-linear interactions between faces are efficiently exploited.
Finally, some wrinkle enhancement together with geometry constraints are incorporated in post-processing steps for more consistent results.
Therefore, plausible synthesized results can be achieved using this technique. 
A comparison in term of synthesis quality between this model and other conventional approaches is shown in Fig. \ref{fig:CompareTRBM_IAAP}.

\begin{figure*}[!t]
	\begin{center}
		\includegraphics[width=17cm]{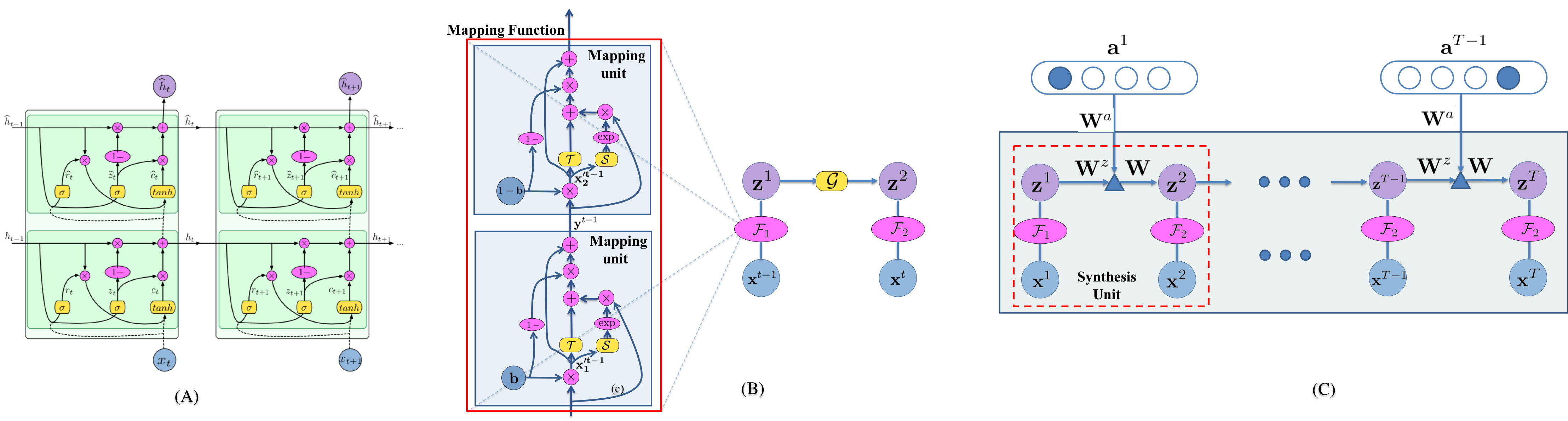}\end{center}
	\caption{Structures of (A) RNN-based model \cite{wang2016recurrent}, (B) Temporal Non-Volume Preserving (TNVP) approach \cite{Duong_2017_ICCV}, and (C) Subject-dependent Deep Aging Path (SDAP) \cite{duong2017learning}. While SDAP shares the Mapping function with TNVP, it aims at embedding the aging transformation of the whole aging sequence.}
	\label{fig:RNN_TNVP_SDAP}
\end{figure*}

\paragraph{\textbf{Recurrent Neural Network-based model}}
Approaching the age progression in a similar way of decomposition, instead of using TRBM, Wang et al. \cite{wang2016recurrent} proposed to use a Recurrent Neural Network with two-layer gated recurrent unit (GRU) to model aging sequence. With the recurrent connection between the hidden units, the model can efficiently exploit  the information from previous faces as \textit{``memory''} to produce smoother transition between faces during synthesizing process. 
Fig. \ref{fig:RNN_TNVP_SDAP}(A) illustrates the architecture of the proposed RNN for age progression. In particular, 
let $\mathbf{x}_t$ be the input face at young age, this network firstly encodes it into latent representation $\mathbf{h}_t$ (hidden/memory units) by the bottom GRU and then decodes this representation into an older face $\mathbf{\hat{h}}_t$ of the subject using the top GRU. The relationship between $\mathbf{x}_t$ and $\mathbf{h}_t$ can be interpreted as follows.
\small
\begin{equation}
\begin{split}
\mathbf{z}_t &= \sigma (\mathbf{W}_{zh} \mathbf{h}_{t-1} + \mathbf{W}_{zx}\mathbf{x}_{t} + \mathbf{b}_z)\\
\mathbf{r}_t &= \sigma (\mathbf{W}_{rh} \mathbf{h}_{t-1} + \mathbf{W}_{rx}\mathbf{x}_{t} + \mathbf{b}_r)\\
\mathbf{c}_t &= \tanh (\mathbf{W}_{ch} \mathbf{r}_{t} \odot \mathbf{h}_{t-1} + \mathbf{W}_{cx}\mathbf{x}_{t} + \mathbf{b}_c)\\
\mathbf{h}_t &= (1-\mathbf{z}_t) \odot \mathbf{h}_{t-1} + \mathbf{z}_t \odot \mathbf{c}_t\\
\end{split}
\end{equation}
\normalsize
Similar formulations are also employed for the relationship between $\mathbf{h}_t$ and $\mathbf{\hat{h}}_t$. Then the difference between $\mathbf{\hat{h}}_t$ and the ground-truth aged face is computed in a form of $\ell_2$ loss function. The system is then trained to obtain the synthesis capability. Finally, in order to generate the wrinkles for the aged-faces, the prototyping-style approach is adopted for wrinkle transferring. Although this approach has produced some improvements comparing to classical approaches, the use of a fixed reconstruction loss function has limited its synthesis ability and usually resulted in blurry  faces.

\paragraph{\textbf{GAN-based model}}
Rather than \textit{step-by-step synthesis} as in previous approaches, Antipov et al. \cite{antipov2017face}, Zhang et al. \cite{Zhang_2017_CVPR}, and Li et al. \cite{li2018global} turned into another direction of age progression, i.e. \textit{direct approach}, and adopted the structure of GAN in their architectures. Fig. \ref{fig:TRBM_GAN}(B) illustrates the structure of the Conditional Adversarial Autoencoder (CAAE) \cite{Zhang_2017_CVPR}. From this figure, one can easily see that the the authors have adopted the GAN structure as presented in Section \ref{sec:GANStructure} with an additional age label feature in the representation of latent variables. By this way, they can further encode the relationship between subject identity related high-level features of the input face and its age label. After training, by simply changing the aging label according to the target age, the deep-neural-network generator is able to synthesize the aged face at that age. Compared to Eqn. \eqref{eqn:GANformulation}, the new objective function is adapted as.
\small
\begin{equation} \nonumber
\begin{split}
\min_{E,G} & \max_{D_z,D_{img}} \lambda \mathcal{L}(\mathbf{x}, G(E(\mathbf{x}),\mathbf{l})) + \gamma TV(G(E(\mathbf{x}),\mathbf{l}))\\
& + \mathbb{E}_{\mathbf{z}^*\sim p(\mathbf{z})}\left[\log D_z(\mathbf{z}^*)\right] + \mathbb{E}_{\mathbf{x} \sim p_{data}(\mathbf{x})} \left[\log(1-D_z(E(\mathbf{x})))\right]\\
& + \mathbb{E}_{\mathbf{x,l} \sim p_{data}(\mathbf{x,l})}\left[\log D_{img}(\mathbf{x,l})\right] \\
& +  \mathbb{E}_{\mathbf{x,l} \sim p_{data}(\mathbf{x,l})}\left[\log(1-D_{img}(G(E(\mathbf{x}),\mathbf{l})))\right]\\
\end{split}
\end{equation}
\normalsize
where $\mathbf{l}$ denotes the vector represented age label; $\mathbf{z}$ is the latent feature vector; $E$ is the decoder function, i.e. $E(\mathbf{x}) = \mathbf{z}$. $\mathcal{L}(\cdot, \cdot)$ and $TV(\cdot)$ are the $\ell_2$ norm and total variation functions, respectively. $p_{data}(\mathbf{x})$ denotes the distribution of the training data. As one can see, the conditional constraint on the age label is represented in the last two terms of the loss function.
Although this model type can avoid the requirement of longitudinal age database during training, it is not easy to be converged due to the step of maintaining a good balance between generator and discriminator which is hard to achieve. Moreover, similar to RNN-based approach, GAN-based models also  incorporate the $\ell_2$-norm in their objective functions. Therefore, their synthesized results are limited in terms of the image sharpness.

\begin{figure*}[!t]
	\begin{center}
		\includegraphics[width=14cm]{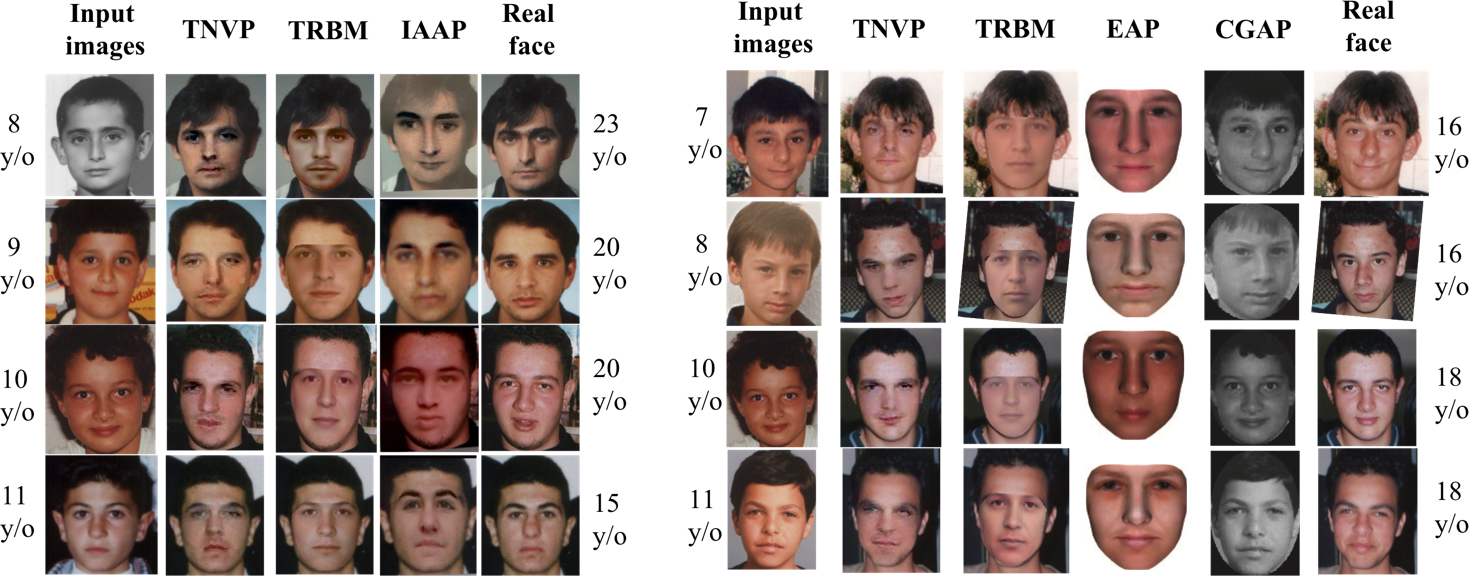}\end{center}
	\caption{A comparison between Deep Learning Approaches, i.e. TNVP \cite{Duong_2017_ICCV}, TRBM \cite{Duong_2016_CVPR}, against conventional approaches including IAAP \cite{kemelmacher2014illumination}, Exemplar based (EAP) \cite{shen2011exemplar}, and Craniofacial Growth (CGAP) \cite{ramanathan2006modeling} models.}
	\label{fig:CompareTNVP_IAAP}
\end{figure*}

\paragraph{\textbf{Temporal Non-Volume Preserving transformation}}
Recently, addressing a limitation of intractable learning process of TRBM based model as well as the image quality of RNN-based and GAN-based approaches,
the Temporal Non-Volume Preserving (TNVP) approach is introduced in \cite{Duong_2017_ICCV} for embedding the feature transformations between faces in consecutive stages while keeping a tractable density function, exact inference and evaluation.
Unlike previous approaches which incorporate only PGM or CNN structures, this proposed model enjoys the advantages of both architectures to improve its image synthesis quality and highly non-linear feature generation.
The idea of this model start from a PGM with relationships between variables in image and latent domains (see Fig. \ref{fig:RNN_TNVP_SDAP}(B)) given by
\small
\begin{equation}
\begin{split}\label{eqn:ModelFormulation}
\mathbf{z}^{t-1}&  = \mathcal{F}_1 (\mathbf{x}^{t-1}; \theta_1)\\
\mathbf{z}^{t} &= \mathcal{H}(\mathbf{z}^{t-1},\mathbf{x}^t; \theta_2, \theta_3)  \\& = \mathcal{G}(\mathbf{z}^{t-1};\theta_3) + \mathcal{F}_2(\mathbf{x}^t;\theta_2)
\end{split}
\end{equation}
\normalsize
where $\mathcal{F}_1, \mathcal{F}_2$ denote the bijection functions mapping $\mathbf{x}^{t-1}$ and $\mathbf{x}^{t}$ to their latent variables $\mathbf{z}^{t-1}, \mathbf{z}^{t}$, respectively. $\mathcal{G}$ is the function embedding the aging transformation between latent variables.
Then the probability density function is derived by.
\small
\begin{equation}
\begin{split} \label{eqn:likelihood}
p_{X^t}(\mathbf{x}^t|\mathbf{x}^{t-1};\theta)&=p_{X^t}(\mathbf{x}^t|\mathbf{z}^{t-1};\theta)\\
&=p_{Z^t}(\mathbf{z}^t|\mathbf{z}^{t-1};\theta)\left|\frac{\partial \mathcal{H}(\mathbf{z}^{t-1}, \mathbf{x}^t;\theta)}{\partial \mathbf{x}^t}\right|\\
&=p_{Z^t}(\mathbf{z}^t|\mathbf{z}^{t-1};\theta)\left|\frac{\partial \mathcal{F}_2(\mathbf{x}^t;\theta)}{\partial \mathbf{x}^t}\right|
\end{split}
\end{equation} 
\normalsize
where $p_{X^t}(\mathbf{x}^t|\mathbf{x}^{t-1};\theta)$ and $p_{Z^t}(\mathbf{z}^t|\mathbf{z}^{t-1};\theta)$ denote the conditional distribution of $\mathbf{x}^t$ and $\mathbf{z}^t$, respectively.
By a specific design of mapping functions $\mathcal{F}_1, \mathcal{F}_2$, the two terms on the right-hand-side of Eqn. \eqref{eqn:likelihood} can be computed exactly and effectively.
As a result, the authors can form a deep CNN network optimized under the concepts of PGM.  While keeping the tractable log-likelihood density estimation in its objective function, the model turns age progression architectures into new direction where the CNN network can avoid using a fix reconstruction loss function and obtain high-quality synthesized faces. Fig. \ref{fig:CompareTNVP_IAAP} illustrates the synthesized results achieved by TNVP in comparison with other approaches.

\paragraph{\textbf{Subject-dependent Deep Aging Path (SDAP) model}}
Inspiring from the advantages of TVNP, the Inverse Reinforcement (IRL) Learning is also taken into account in the structure of Subject-dependent Deep Aging Path (SDAP) model  \cite{duong2017learning}. 
Under the hypothesis that each subject should have his/her own facial development, Duong et al. \cite{duong2017learning} proposed to use an additional aging controller in the structure of TNVP. Then rather than only embedding the aging transformation between pairwise relationship between consecutive age groups, the SDAP structure learns from the aging transformation of the whole face sequence for better long-term aging synthesis. This goal is achieved via a Subject-Dependent Aging Policy Network which guarantees to provide an appropriate planning aging path for the age controller corresponding to the subject's features. The most interesting point of SDAP is that this is one of the pioneers incorporating IRL framework into age progression task.
In this approach, let $\zeta_i = \{\mathbf{x}^1_i, \mathbf{a}^1_i, \ldots, \mathbf{x}^T_i\}$ be the age sequence of $i$-th subject where $ \{\mathbf{x}^1_i, \ldots, \mathbf{x}^T_i\}$ are the face sequence representing the facial development of $i$-th subject  and $ \mathbf{a}^j_i $ denote the variables control the aging amount added to $ \mathbf{x}^j_i $ to become $ \mathbf{x}^{j+1}_i $. The probability of $\zeta_i$  can be formulated via an energy function $E_\Gamma(\zeta_i)$  by
\small
\begin{equation} \label{eqn:ProbabilitySDAP}
P(\zeta_i) = \frac{1}{Z} \exp(-E_\Gamma(\zeta_i))
\end{equation}
\normalsize
where $Z$ is the partition function. Notice that the formulation of Eqn. \eqref{eqn:ProbabilitySDAP} is very similar to joint distribution between variables of RBM as in Eqn. \eqref{eqn:RBM}.
Then the goal is to learn a Subject-Dependent Aging Policy Network that can predict $ \mathbf{a}^j_i $ for each $ \mathbf{x}^j_i $ during synthesized process. The objective function is defined as.
\small
\begin{equation} 
\label{eqn::Log_likelihood_sequence}
\begin{split}
\Gamma^* &= \arg \max_\Gamma \mathcal{L} (\zeta;\Gamma)=\frac{1}{M} \log \prod_{\zeta_i \in \zeta} P(\zeta_i)
\end{split}
\end{equation}
\normalsize
\begin{figure}[!t]
	\begin{center}
		\includegraphics[width=7.2cm]{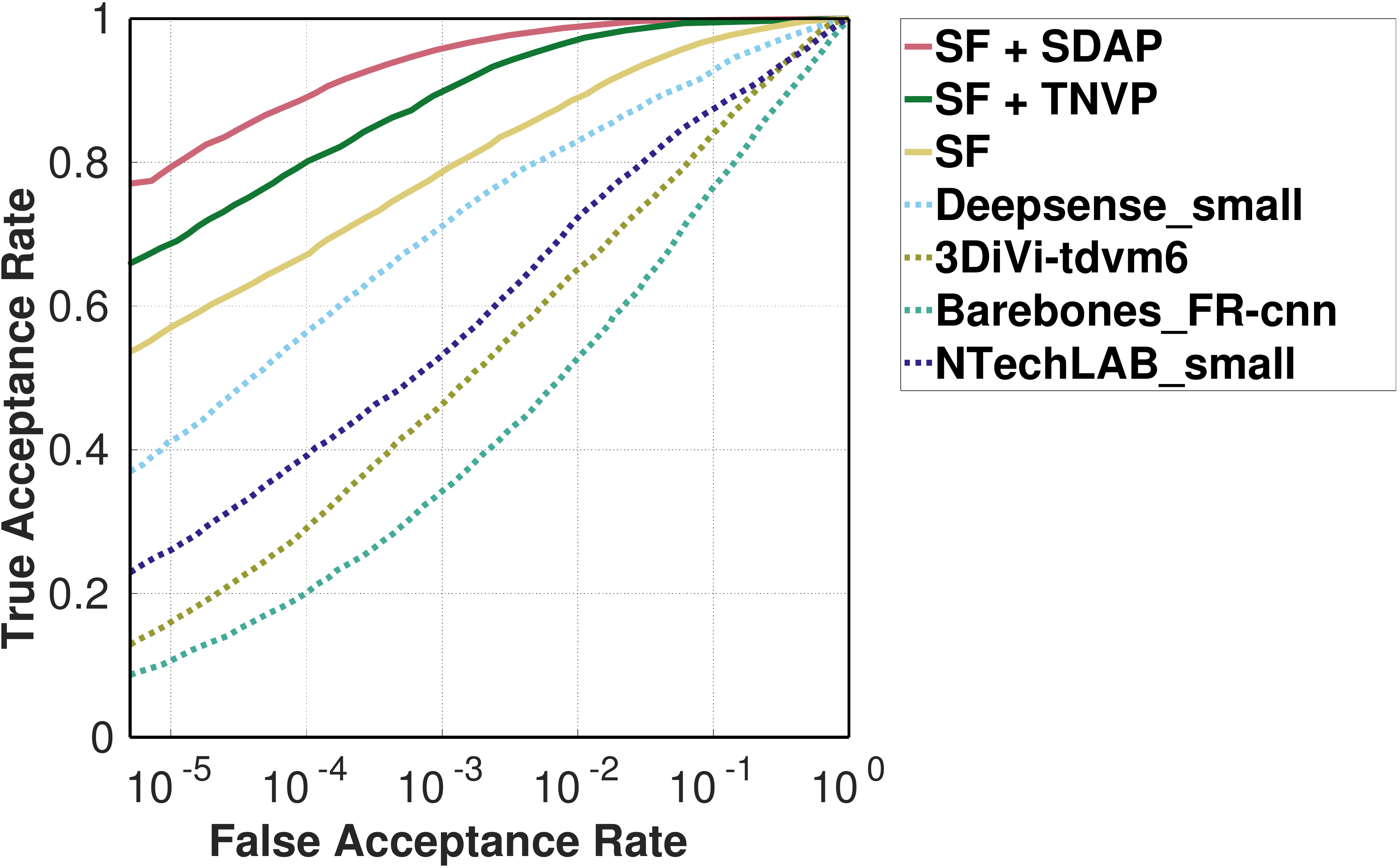}\end{center}
	\caption{An example of age invariant face recognition by using age progression models, i.e. TVNP \cite{Duong_2017_ICCV} and SDAP \cite{duong2017learning}. By incorporated their synthesized results, the accuracy of face recognition (FR) system can be improved significantly. Note that the results of other FR methods are provided in Megaface website\cite{kemelmacher2016megaface}.}
	\label{fig:CompareRecognition}
\end{figure}
Finally, a specific design of IRL framework is proposed to learn the Policy Network.
From the experimental results, SDAP has shown its potential to outperform TNVP and other approaches on synthesis results and cross-age verification accuracy. As shown in Fig. \ref{fig:CompareRecognition}, SDAP can help to significantly improve the accuracy for face recognition system.

\section{Conclusion}
In this paper, we have reviewed the main structures of Deep Generative Models for Age Progression task.  Compared to other classical approaches, 
Deep Learning has shown its potential either in learning the highly non-linear age variation or aging transformation embedding. As a result, not only do their synthesized faces improve in the image quality but also help to significantly boost the recognition accuracy for cross-age face verification system.
Several common aging databases that support the facial modeling and aging embedding process are also discussed.

\bibliographystyle{ieee}
\bibliography{egbib}

\end{document}